\documentclass[10pt, twoside]{article} 

\usepackage[utf8]{inputenc}
\usepackage[T1]{fontenc}   
\usepackage{longtable}
\usepackage{url}           
\usepackage{booktabs}      
\usepackage{amsfonts}      
\usepackage{nicefrac}      
\usepackage{microtype}     
\usepackage{lipsum}        
\usepackage{graphicx}
\usepackage[numbers]{natbib}
\usepackage{amsmath}
\usepackage{amsthm}
\usepackage{algorithm}
\usepackage{algorithmic}
\usepackage{epsf}
\usepackage{epsfig}
\usepackage{fancyhdr}
\usepackage{graphics}
\usepackage{graphicx}
\usepackage{psfrag}
\usepackage{fullpage}
\usepackage{pdfpages}
\usepackage{color}
\usepackage{amssymb,bbm}
\usepackage{caption}
\usepackage{textcomp}
\usepackage{siunitx}
\usepackage{wrapfig}
\usepackage{multirow}
\usepackage{multicol}
\usepackage[colorlinks,linkcolor=magenta,citecolor=blue, pagebackref=true]{hyperref}
\usepackage{gensymb}
\usepackage{enumitem}
\usepackage{subfig}
\usepackage{float}
\usepackage{wrapfig}
\usepackage{tabularx}
\usepackage{colortbl}
\usepackage{xcolor}
\usepackage{tikz}
\usepackage{authblk}

\usetikzlibrary{shapes.geometric, arrows, positioning}
\tikzset{
    block/.style = {rectangle, draw, text width=5em, text centered, rounded corners, minimum height=4em},
    line/.style = {draw, -latex'}
}
\definecolor{darkgreen}{rgb}{0.0, 0.5, 0.0}
\usepackage{esvect}

\title{Equivariant vs. Invariant Layers: A Comparison of Backbone and Pooling for Point Cloud Classification}

\author[1]{Abihith Kothapalli}
\author[1]{Ashkan Shahbazi}
\author[1]{Xinran Liu}
\author[1]{Robert Sheng}
\author[1]{Soheil Kolouri}
\affil[1]{Department of Computer Science, Vanderbilt University}
\affil[1]{\{abi.kothapalli, ashkan.shahbazi, xinran.liu, robert.sheng, soheil.kolouri\}@vanderbilt.edu}

\date{}

\begin{document}

\maketitle

\begin{abstract}
Learning from set-structured data, such as point clouds, has gained significant attention from the machine learning community. Geometric deep learning provides a blueprint for designing effective set neural networks that preserve the permutation symmetry of set-structured data. Of our interest are permutation invariant networks, which are composed of a permutation equivariant backbone, permutation invariant global pooling, and regression/classification head. While existing literature has focused on improving equivariant backbones, the impact of the pooling layer is often overlooked. In this paper, we examine the interplay between permutation equivariant backbones and permutation invariant global pooling on three benchmark point cloud classification datasets. Our findings reveal that: 1) complex pooling methods, such as transport-based or attention-based poolings, can significantly boost the performance of simple backbones, but the benefits diminish for more complex backbones, 2) even complex backbones can benefit from pooling layers in low data scenarios, 3) surprisingly, the choice of pooling layers can have a more significant impact on the model's performance than adjusting the width and depth of the backbone, and 4) pairwise combination of pooling layers can significantly improve the performance of a fixed backbone. Our comprehensive study provides insights for practitioners to design better permutation invariant set neural networks. Our code is available at \url{https://github.com/mint-vu/backbone_vs_pooling}.
\end{abstract}

\section{Introduction}

Set classification is a challenging problem in machine learning that has numerous real-world applications, including computer vision \cite{hu2012face,lu2015multi, Wang_2015_CVPR, Ye_2020_CVPR}, natural language processing \cite{bahdanau2014neural, liu2018learning, gong2020hierarchical, NEURIPS2021_3bbca1d2}, and bioinformatics \cite{skianis2020rep, NEURIPS2021_bd4a6d05, kim2022differentiable}. Recently, there have been significant developments in 3D data acquisition, leading to increased interest in representation learning from point cloud data, which consists of sets of unordered 3D points. However, learning from point clouds presents inherent challenges, such as noise, occlusion, and irregularity, which can make representation learning difficult \cite{Xiang_2021_ICCV, Qi_2017_CVPR, hua2018pointwise, liu2019point2sequence, NIPS2017_d8bf84be, Prokudin_2019_ICCV, pmlr-v80-achlioptas18a, Hassani_2019_ICCV, ravanbakhsh2017deep}.

\begin{figure*}[t!]
  \centering
  \includegraphics[width=\linewidth]{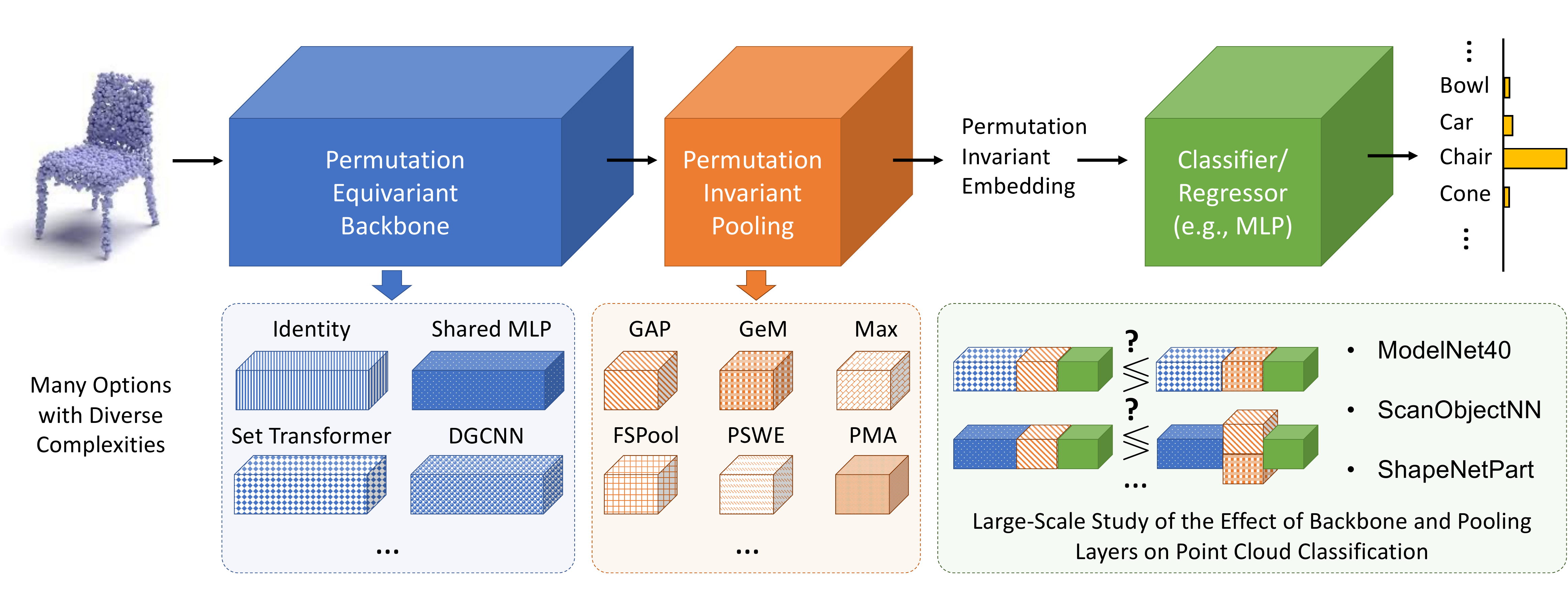}
  \vspace{-.3in}
  \caption{\textbf{Investigating the Impact of Backbone and Pooling Combinations on Point Cloud Data.} Through comprehensive experiments on three point cloud benchmarks, we evaluate the performance of models using different combinations of permutation equivariant backbones with permutation invariant pooling techniques. Additionally, we demonstrate the effectiveness of combining specific poolings to improve model performance. Our study provides insights into the benefits of equivariant and invariant layers for point cloud analysis.}
  \label{fig:overall}
\end{figure*}

Most existing methods for point cloud classification can be grouped into three categories: voxel-based, projection-based, and point-based methods. 
Voxel-based methods are volumetric approaches that rasterize point clouds onto a three-dimensional grid and utilize 3D convolutional neural networks (CNNs) to learn from the rasterized data \cite{ben20183dmfv,hermosilla2018monte,maturana2015voxnet,hua2018pointwise,Mao_2019_ICCV,Zhang_2019_ICCV}. Projection-based approaches project/render the point cloud onto 2D images (e.g., depth maps) from multiple orthogonal views and then process these images with 2D CNNs \cite{kanezaki2018rotationnet, yu2018multi, Su_2015_ICCV, 9506426,goyal2021revisiting,NEURIPS2018_f5f8590c}. Lastly, point-based methods involve permutation invariant neural networks based on the principles of geometric deep learning \cite{bronstein2021geometric}. These methods \cite{Xiang_2021_ICCV, dgcnn, Qi_2017_CVPR, lee2019set, li2018so, shen2018mining, dominguez2018general, liu2019point2sequence} utilize permutation invariant functions to exploit the inherent permutation symmetry in point clouds. They often involve using a permutation equivariant (or semi-equivariant) backbone, followed by a permutation invariant global pooling to obtain a permutation invariant representation that is then passed into a classifier/regressor. This paper focuses on point-based approaches for point cloud classification rooted in geometric deep learning.

The majority of existing research in point cloud classification has focused on designing novel neural architectures, specifically permutation equivariant (or semi-equivariant) backbones \cite{dgcnn, Qi_2017_CVPR, zaheer2017deep, Xiang_2021_ICCV, lee2019set}, while some recent work has highlighted the significance of pooling layers \cite{murphy2018janossy, zhang2020fspool, naderializadeh2021PSWE, e24121745, lee2019set, 9053217,skianis2020rep,mialon2021trainable}. Our objective in this paper is to investigate the intricate interplay between backbone architecture and pooling approach on model performance in point cloud classification. We conduct extensive studies on three prominent point cloud classification benchmark datasets --- ModelNet40 \cite{wu20153d}, ScanObjectNN \cite{uy-scanobjectnn-iccv19}, and ShapeNetPart \cite{chang2015shapenet} --- to explore the impact of model architectural choices and complexities on performance.

In our experiments, we aimed to investigate the interplay between the choice of permutation equivariant backbones and permutation invariant pooling approaches on point cloud classification performance. To this end, we evaluated seven different permutation equivariant backbones: Identity; DeepSets \cite{zaheer2017deep}; (Induced) Set Attention Blocks (SAB and ISAB) \cite{lee2019set}; Dynamic Graph CNN (DGCNN) \cite{dgcnn}; PointNet \cite{Qi_2017_CVPR}; and CurveNet \cite{Xiang_2021_ICCV}. For poolings, we consider eleven different permutation invariant pooling approaches: Global Average Pool (GAP); Generalized Mean (GeM) \cite{radenovic2018fine}; Max and k-Max \cite{kim2014convolutional}; Pooling by Multihead Attention (PMA) \cite{lee2019set}; Global Multi-Head Attentive (GMHA) and Multi-Resolution Multi-Head Attentive (MMHA) poolings \cite{9053217}; Pooling by Sliced-Wasserstein Embedding (PSWE) \cite{naderializadeh2021PSWE} (both learnable and non-learnable versions); Featurewise Sort Pooling (FSPool) \cite{zhang2020fspool}; and RepSet \cite{skianis2020rep}. While there have been many recent advancements in point cloud classification and segmentation models beyond what we discuss here, the purpose of this work is to focus only on permutation equivariant backbones and permutation invariant pooling layers, and hence many of these recent advancements fall outside the scope of this work. To ensure fair comparisons and avoid the effects of auxiliary factors like evaluation schemes, data augmentation strategies, and loss functions, we devised a unified training and evaluation scheme. We evaluated the performance of all pairs of backbones and pooling methods (a total of 77 models) on three benchmark datasets, namely ModelNet40 \cite{wu20153d}, ScanObjectNN \cite{uy-scanobjectnn-iccv19}, and ShapeNetPart \cite{chang2015shapenet}. We also investigated the effect of using multiple pooling layers for fixed backbones and provided the performance of the 77 models under full and restricted training data.

Our findings can be summarized as follows. First, we observed a performance gap between traditional permutation invariant pooling approaches and more recent OT-based and attention-based pooling approaches. This gap is most prominent with simpler permutation equivariant backbones and becomes less significant as the backbone's complexity increases. Additionally, we demonstrated that this performance gap is more pronounced when working with limited amounts of data. Second, we discovered that OT-based methods exhibit less sensitivity to the sample size, providing stable performance even in low-data scenarios, whereas attention-based methods exhibit more variation in their performance. Third, we observed that pooling layers play a more critical role in performance of DeepSets \cite{zaheer2017deep} and Set Transformers \cite{lee2019set} compared to the width and depth of the backbone. Lastly, we observed that certain permutation invariant pooling layers are complementary, and combining them can significantly enhance the model's performance.

{\bf Contributions.} We demonstrate that: a) Permutation invariant pooling layers play a crucial role in point cloud classification.
b) Transport-based pooling layers exhibit less sensitivity to training data size. Models with simple and shallow permutation equivariant backbones, combined with advanced pooling techniques, perform well, particularly in low-data scenarios.
c) In the context of DeepSets \cite{zaheer2017deep} and Set Transformers \cite{lee2019set}, the choice of the pooling layer has a greater impact on performance than adjustments in backbone depth and width.
d) Pairing specific pooling layers can result in significant performance improvements.
Overall, our study provides valuable insights for the community in designing effective permutation invariant models and raises further awareness on the importance of global pooling layers.


\section{Related Work}

Numerous recent studies have been dedicated to addressing point set and graph classification problems by devising neural architectures that exhibit permutation invariance. These architectures typically consist of multiple permutation equivariant layers coupled with permutation invariant pooling operations. In the context of our focus on point cloud classification, here we provide a concise summary of recent advancements in developing permutation equivariant backbones and invariant pooling techniques. We also note that this work aims to focus only on permutation equivariant backbones and permutation invariant pooling layers; in this regard, while there have been many recent advancements in point cloud classification and segmentation models, even outperforming some of the methods we consider here, these models are not always permutation equivariant, and hence we do not consider them in this work.

\textbf{Permutation Equivariant Backbones.} In point clouds, the input features for each point typically consist of coordinates, occasionally supplemented by a surface normal. However, these input features alone often lack sufficient descriptions of shape geometries, rendering them insufficient for classification purposes. To address this limitation, permutation equivariant backbones play a crucial role by enabling the aggregation and enrichment of these features for each point. This process results in enhanced features encompassing local and global information, providing a more comprehensive shape description. PointNet \cite{Qi_2017_CVPR} and its subsequent extension, PointNet++ \cite{NIPS2017_d8bf84be}, have emerged as pioneering network architectures for point cloud classification. PointNet utilizes permutation equivariant blocks, which consist of a spatial/feature transformer network \cite{jaderberg2015spatial}, followed by a shared multi-layer perceptron (MLP) and featurewise max-pooling operations. These computational blocks in PointNet closely resemble the backbone of the DeepSets architecture \cite{zaheer2017deep}, wherein a shared MLP is applied to the set elements to generate a permutation equivariant backbone.

Shared MLP-based permutation equivariant backbones, while widely used, have inherent limitations in their representation capabilities \cite{lee2019set}, as they lack the ability to facilitate interactions or message passing between neighboring points. Conversely, alternative methods such as PointCNN \cite{NEURIPS2018_f5f8590c}, Set Transformers \cite{lee2019set}, Dynamic Graph CNN (DGCNN) \cite{dgcnn}, and CurveNet \cite{Xiang_2021_ICCV} offer more flexible permutation equivariant backbones by enabling message passing between set elements. In particular, the Set Transformer framework, examples of which include the Set Attention Block (SAB) and Induced Set Attention Block (ISAB) methods \cite{lee2019set}, leverages the powerful self-attention mechanism \cite{vaswani2017attention} and its variations as the computational building blocks for the backbone. Similarly, DGCNN \cite{dgcnn} modifies the message passing between input points by dynamically constructing a graph in each layer and utilizing edge convolutions. Lastly, CurveNet \cite{Xiang_2021_ICCV} employs a guided walk on the point cloud to identify curve groupings, enabling more informative message passing within the network architecture. 

\textbf{Permutation Invariant Poolings.} Permutation invariant pooling layers play a pivotal role in geometric deep learning, where the sum, average, and max/min functions are widely acknowledged as the simplest and commonly used permutation invariant functions in the literature. However, recent advancements have introduced more sophisticated permutation invariant pooling layers that offer enhanced performance. This emerging work stems from the observation that traditional pooling layers, such as Global Average Pooling (GAP), may not adequately capture the feature distribution extracted by the permutation equivariant backbone. To address this, researchers have proposed novel approaches such as covariance pooling, introduced by \cite{acharya2018covariance} and \cite{wang2020deep}, which aims to capture the second moment of the feature distribution. Similarly, \cite{radenovic2018fine} proposed Generalized Mean (GeM) pooling, which approximates higher moments of the feature distribution while mitigating the computational cost associated with covariance pooling. \cite{murphy2018janossy} introduced Janossy pooling, which represents permutation invariant functions as the average of ``permutation-sensitive'' functions applied to all reorderings of the input sequence. Furthermore, recent studies have explored permutation invariant pooling layers based on optimal transport (OT) theory, extending beyond the second moment of distributions. For example, \cite{kolouri2021wasserstein} and \cite{mialon2021trainable} concurrently introduced permutation invariant pooling layers for graph neural networks using Wasserstein embedding (also known as linearized OT \cite{wang2013linear,moosmuller2023linear}). \cite{naderializadeh2021PSWE} expanded upon the Wasserstein embedding framework with their proposal of Pooling by Sliced Wasserstein Embedding (PSWE), a suitable permutation invariant function for end-to-end learning that encodes the backbone's feature distribution. Notably, \cite{zhang2020fspool} had previously introduced Featurewise Sort Pooling (FSPool), which can be considered a specific case of PSWE. \cite{skianis2020rep} developed RepSet and its approximation version, ApprRepSet, which leverages bipartite matching and is similar to OT-based pooling layers.

Attention-based permutation invariant pooling layers have emerged as another family of pooling layers and have gained significant attention in geometric deep learning due to their ability to capture complex relationships and dependencies among features while maintaining permutation invariance. These pooling layers leverage the attention mechanisms \cite{vaswani2017attention}, allowing the model to assign weights to different features based on their relevance dynamically. The attention mechanism enables the pooling layer to aggregate important features while suppressing less significant ones, thus enhancing the overall representation power of the layer. Several studies have proposed attention-based permutation invariant pooling layers for geometric deep learning tasks. For example, \cite{lee2019self} presented Self-Attention Graph Pooling, where attention coefficients are learned to guide the pooling process based on node-level features. Similarly, \cite{e24121745} introduced graph multi-head pooling which utilizes attention as its core pooling mechanism. \cite{lee2019set} introduced pooling by multi-head attention (PMA) and \cite{9053217} proposed global multi-head attentive (GMHA) and multi-resolution multi-head attentive (MMHA) poolings. These attention-based pooling layers not only improve the model's ability to capture important features but also provide a mechanism to adaptively aggregate information from different nodes or regions in a permutation invariant manner.

Although significant progress has been made in developing novel permutation equivariant backbones and permutation invariant pooling layers in geometric deep learning, the interaction between these two modules has not been thoroughly investigated in prior works. In our paper, we aim to bridge this gap by conducting an extensive study on the interplay between the backbone and pooling layers on three benchmark point cloud classification datasets. By analyzing the performance of different combinations of backbones and pooling layers, we provide valuable insights into designing effective and efficient geometric deep learning models. Our study aims to shed light on the role of these modules and their interaction in point cloud classification tasks.

\section{Experiments}
\label{sec:experiment}

Our study encompasses a comprehensive set of experiments aimed at addressing key questions in the context of point cloud classification. Firstly, we investigated the impact of combining various permutation invariant pooling techniques with permutation equivariant backbone architectures on classification performance. Secondly, we explored how this combination affects learning when dealing with limited training data. Thirdly, we analyzed the effects of adjusting the depth and width of a backbone, specifically DeepSets and Set Transformer (in particular, the SAB method), on the network's performance when different pooling layers are utilized. Lastly, we examined the benefits of pairing specific pooling layers with a given backbone architecture. 

We conducted experiments on three point cloud classification benchmarks: ModelNet40 \cite{wu20153d}, ScanObjectNN \cite{uy-scanobjectnn-iccv19}, and ShapeNetPart, which is the point cloud variant of the ShapeNet dataset \cite{chang2015shapenet}. To avoid nuisance training variations, like different types of augmentations, different loss functions, training hyperparameters, etc., we designed a unified experimental setup for all datasets and for all models. In what follows, we first provide a brief description of the datasets, then review the permutation equivariant backbones followed by the different pooling layers and the classifier head we used in this work. 

\subsection{Datasets} 

\textbf{ModelNet40} \cite{wu20153d} is a popular benchmark in 3D object recognition, comprising 12,311 CAD models from 40 object categories, with around 300 instances per category. This dataset offers diverse objects in terms of shape and size, making it well-suited for evaluating the performance of 3D point cloud classification models. Each instance in the dataset consists of 2048 points.

\textbf{ScanObjectNN} \cite{uy-scanobjectnn-iccv19} is a real-world point cloud object dataset derived from scanned indoor scenes. It encompasses nearly 15,000 objects across 15 categories, with 2902 unique object instances. In our experiments, we focus solely on utilizing the global coordinates of each point per sample, disregarding other attributes such as normals, color attributes, semantic labels, and part annotations provided by the original ScanObjectNN dataset. Like ModelNet40, each sample in ScanObjectNN comprises 2048 points.

\textbf{ShapeNetPart} is the point cloud counterpart of the ShapeNet \cite{chang2015shapenet} dataset, comprising more than 15,000 3D models across 16 object categories, including furniture, cars, airplanes, and animals. The objects are originally represented as 3D meshes, but here we consider the mesh nodes as a 3D point cloud, disregarding the edges. Each model in this dataset contains a variable number of points, typically ranging from approximately 500 to 3000 points for a given object.

For all datasets, we randomly sampled 1024 points for each object to construct a batch. For the ModelNet40 and ScanObjectNN datasets, we sampled these points without replacement, and for the ShapeNetPart dataset, we sampled these points with replacement (as there may be fewer than 1024 points initially). Furthermore, we followed the work of \cite{naderializadeh2021PSWE} and applied random translation, rotation, and jittering for data augmentation in all experiments. 

\subsection{Model Blueprint}

\textbf{Permutation Equivariant Backbones.} In our experiments, we carefully selected diverse backbones with varying forward time, backward time, and model size. To establish a baseline, we utilized the Identity backbone (i.e., lack of a backbone) with different pooling techniques. Subsequently, we conducted similar experiments on several other backbones, including MLP (as used in DeepSets), SAB \cite{lee2019set}, ISAB \cite{lee2019set}, DGCNN \cite{dgcnn}, PointNet \cite{Qi_2017_CVPR}, and CurveNet \cite{Xiang_2021_ICCV}. For detailed information regarding the hyperparameters of the backbones, we refer readers to Appendix \ref{appendix:backbone}.

\textbf{Permutation Invariant Global Pooling.}  We explored the role of different common and state-of-the-art pooling methods in learning from sets. Our experiments include the classic pooling techniques, including GAP, Max, k-Max \cite{kim2014convolutional}, and GeM \cite{radenovic2018fine}, as well as more complex pooling methods like Learnable and Frozen (Non-Learnable) versions of PSWE \cite{naderializadeh2021PSWE} (which we denote L-PSWE and F-PSWE, respectively), FSPool \cite{zhang2020fspool}, and ApproxRepSet \cite{skianis2020rep}. We also considered attention-based pooling mechanisms, such as PMA \cite{lee2019set}, GMHA \cite{9053217}, and MMHA \cite{9053217}, on different backbones. Further details on pooling hyperparameters can be found in Appendix \ref{appendix:pool}.

\textbf{Classifier Head.} For all models, our classifier consists of a feed-forward network with three hidden layers, each consisting of 128 nodes, and a final layer with its width equal to the number of classes in the dataset. We apply batch normalization and leaky ReLU activation to each hidden layer, as well as dropout after each hidden layer. 


\subsection{Training and Evaluation Settings}
In our experiments, we allocated 1\% of the training data as the validation set. For each model, we simultaneously trained the backbone, pooling, and classifier parameters using the Adam optimizer \cite{diederik2014adam} with cross entropy loss \cite{ackley1985learning} and batches of 32 samples. The initial learning rate was set to $8 \times 10^{-4}$, and we applied a learning rate decay of $\gamma = 0.5$ every 50 epochs for all experiments. Early stopping was considered when the validation loss did not improve for 20 consecutive epochs, and training was capped at 500 epochs if early stopping was not triggered. We used classification accuracy on the test set as our evaluation criterion. Each experiment was conducted with different random seeds three times, and we report the average performance per model. We utilized GPU clusters with 16 NVIDIA RTX A6000 GPUs, AMD EPYC 7713 64-Core Processor, and 256GB of memory. 

\section{Results}

\begin{table}[h!]%
\centering
\setlength{\tabcolsep}{3pt}
\begin{tabular}{c|c|ccccccc|c}
\toprule

&\textbf{Models} &   \textbf{Id} &   \textbf{MLP} &  \textbf{SAB} &  \textbf{ISAB} &     \textbf{DGCNN} &    \textbf{PN} &     \textbf{CN} &     \textbf{P Avg.}\\
\cmidrule{2-10}
\multirow{12}{*}{{\rotatebox[origin=c]{90}{\textbf{ModelNet40} }}}
& GAP       &   4.01 &  40.02 &  83.81 &  83.35 &  85.80 &   84.67 &  89.93 & 67.37\\
& GeM &  26.52 &  38.74 &  85.74 &  83.96 &  89.76 &   85.86 &   \textbf{90.37} & 71.56\\
&Max              &  28.46 &  31.77 &  \textbf{85.86} &  \textbf{85.80} &  89.72 &   \textbf{86.93} &   88.82 & 71.05\\
&k-Max           &  27.76 &  30.25 &  84.70 &  83.48 &  89.94 &   \textbf{87.26} &   90.18 & 70.51\\
&PMA              &  35.20 &  48.96 &  75.96 &  63.42 &  87.99 &   85.84 &   88.78 & 69.45\\
&F-PSWE     &  \textbf{80.57} &  \textbf{80.25} &  83.97 &  \textbf{85.46} &  \textbf{90.65} &   \textbf{86.98} &   89.50 & \textbf{85.34}\\
&L-PSWE   &  \textbf{78.61} &  \textbf{81.65} &  82.53 &  83.19 &  \textbf{90.40} &   86.39 &   \textbf{90.79} & 84.79\\
&FSPool          &  39.93 &  41.02 &  \textbf{86.22} &  83.57 &  \textbf{90.64} &   84.88 &   89.54 & 73.69\\
&GMHA           &  \textbf{78.50} &  75.12 &  79.93 &  \textbf{85.62} &  90.38 &   85.32 &   90.34 & 83.60\\
&MMHA            &  78.31 &  \textbf{78.33} &  \textbf{85.82} &  84.79 &  89.41 &   86.62 &   \textbf{90.85} & 84.87\\
&ApprRepSet       &  29.83 &  39.62 &  84.46 &  84.99 &  89.53 &   84.95 &   89.97 & 71.91\\
\cmidrule{2-10}
&\textbf{B Avg.} & 46.16   &	53.25  &  83.54  &	82.51  &	89.48  &	85.97 &	\textbf{89.92} & \\
\toprule
\multirow{12}{*}{{\rotatebox[origin=c]{90}{\textbf{ScanObjectNN}}}}
& GAP          & 13.39 & 33.34 & 66.17 & 60.78 & 76.41 & 52.71 & 73.18 &    53.71\\
&GeM    & 38.12 & 37.02 & 65.11 & 62.64 & 73.56 & 60.44 & 73.83 &    58.67\\
&Max                 & 30.67 & 36.68 & \textbf{66.38} & \textbf{66.04} & \textbf{77.88} & \textbf{62.33} & \textbf{77.64} &    59.66\\
&k-Max               & 31.32 & 33.96 & 65.49 & \textbf{64.32} & 73.70 & 61.74 & \textbf{77.47} &    58.29\\
&PMA                 & 38.26 & 34.92 & 53.43 & 42.58 & 73.21 & 55.32 & \textbf{78.50} &    53.75\\
&F-PSWE        & \textbf{51.92} & \textbf{56.35} & 65.28 & 63.19 & \textbf{76.65} & \textbf{63.08} & 76.99 &    \textbf{64.78}\\
&L-PSWE      & \textbf{50.27} & \textbf{55.36} & 66.21 & 62.71 & 76.24 & 51.99 & 75.86 &    62.66\\
&FSPool             & 38.77 & 40.08 & 66.07 & \textbf{63.56} & 73.18 & 55.43 & 74.38 &    58.78\\
&GMHA               & \textbf{48.94} & \textbf{52.13} & \textbf{66.69} & 60.75 & \textbf{76.99} & \textbf{63.19} & 76.10 &    63.54\\
&MMHA                & 45.95 & 47.84 & 62.81 & 60.85 & 70.05 & 61.50 & 77.27 &    60.90\\
&ApprRepSet          & 37.19 & 33.83 & \textbf{66.96} & 61.54 & 68.65 & 60.30 & 75.69 &    57.74\\
\cmidrule{2-10}
&\textbf{B Avg.}     & 38.62 & 41.95 & 64.60 & 60.81 & 74.23 & 58.91 & \textbf{76.08} & \\
\toprule
\multirow{12}{*}{{\rotatebox[origin=c]{90}{\textbf{ShapeNetPart}}}} & GAP       &  29.11 &  88.33 &  97.35 &  86.09 &  97.81 &   \textbf{97.85} &   98.26 & 84.97\\

&GeM &  70.21 &  87.59 &  97.14 &  94.44 &  98.37 &   97.35 &   98.19 & 91.90\\

&Max              &  66.34 &  79.59 &  97.14 &  \textbf{97.39} &  98.37 &   \textbf{97.91} &   97.53 & 90.61\\
&k-Max          &  65.92 &  80.96 &  97.62 &  96.76 &  98.37 &   97.25 &   \textbf{98.44} & 90.76\\
&PMA           &  64.66 &  93.52 &  97.36 &  \textbf{97.83} &  98.06 &   97.01 &   98.33 & 92.39\\
&F-PSWE      &  \textbf{96.35} &  \textbf{97.25} &  \textbf{98.05} &  94.09 &  98.44 &   97.46 &   \textbf{98.47} & 97.16\\
&L-PSWE   &  \textbf{96.49} &  \textbf{96.13} &  \textbf{97.79} &  95.75 &  \textbf{98.51} &   97.53 &   98.02 & \textbf{97.17}\\
&FSPool         &  83.22 &  92.58 &  97.60 &  \textbf{96.80} &  98.40 &   \textbf{97.74} &   98.26 & 94.94\\
&GMHA            &  \textbf{95.16} &  94.95 &  97.53 &  96.52 &  98.01 &   97.42 &   \textbf{98.47} & 96.87\\
&MMHA             &  94.54 &  \textbf{95.79} &  \textbf{98.15} &  96.52 &  \textbf{98.61} &   97.18 &   98.40 & 97.03\\
&ApprRepSet      &  82.23 &  85.37 &  97.35 &  96.23 &  \textbf{98.58} &   97.47 &   98.37 & 93.66\\
\cmidrule{2-10}
&\textbf{B Avg.}  &  76.75   & 90.19    &	97.55 &	95.31 &	\textbf{98.32} &	97.47 &	98.25 &\\
\bottomrule
\end{tabular}

\vspace{-.1in}
\caption{We present the results of 77 models across three datasets, highlighting the top three performers for each backbone and the best-performing pooling layer and backbone on average. Here, ``Id'' refers to identity, ``MLP'' refers to the shared MLP backbone in DeepSets, ``PN'' refers to PointNet, and ``CN'' refers to CurveNet.}
\label{tab:table(1)}
\end{table}

\clearpage
\textbf{Pooling vs. Backbone for Point Cloud Classification.} We conducted experiments on the pairwise combination of seven permutation equivariant backbones and eleven permutation invariant pooling layers, as described in Section \ref{sec:experiment}. Each model was trained on the three benchmark datasets, and we repeated each experiment three times, resulting in 231 trained models per dataset. The average test accuracy of the models is reported in Table \ref{tab:table(1)}. Additionally, we provide the average performance of each backbone across all pooling techniques and the average performance of each pooling technique across all backbones. The top three pooling layers for each backbone and the best overall backbone and pooling methods are shown in bold. 

In our experiments, where we controlled for all other parameters, the impact of pooling layers on performance became evident. The OT-based pooling methods, specifically F-PSWE, consistently achieved excellent results across different backbones. The attention-based pooling layers also demonstrated impressive performance. Intriguingly, we observed that both the OT-based and attention-based pooling techniques yielded satisfactory performance even in the absence of a backbone (i.e., with an identity backbone), highlighting the modeling flexibility of these methods.

\begin{figure*}[h!] 
    \centering  \includegraphics[width=\linewidth]{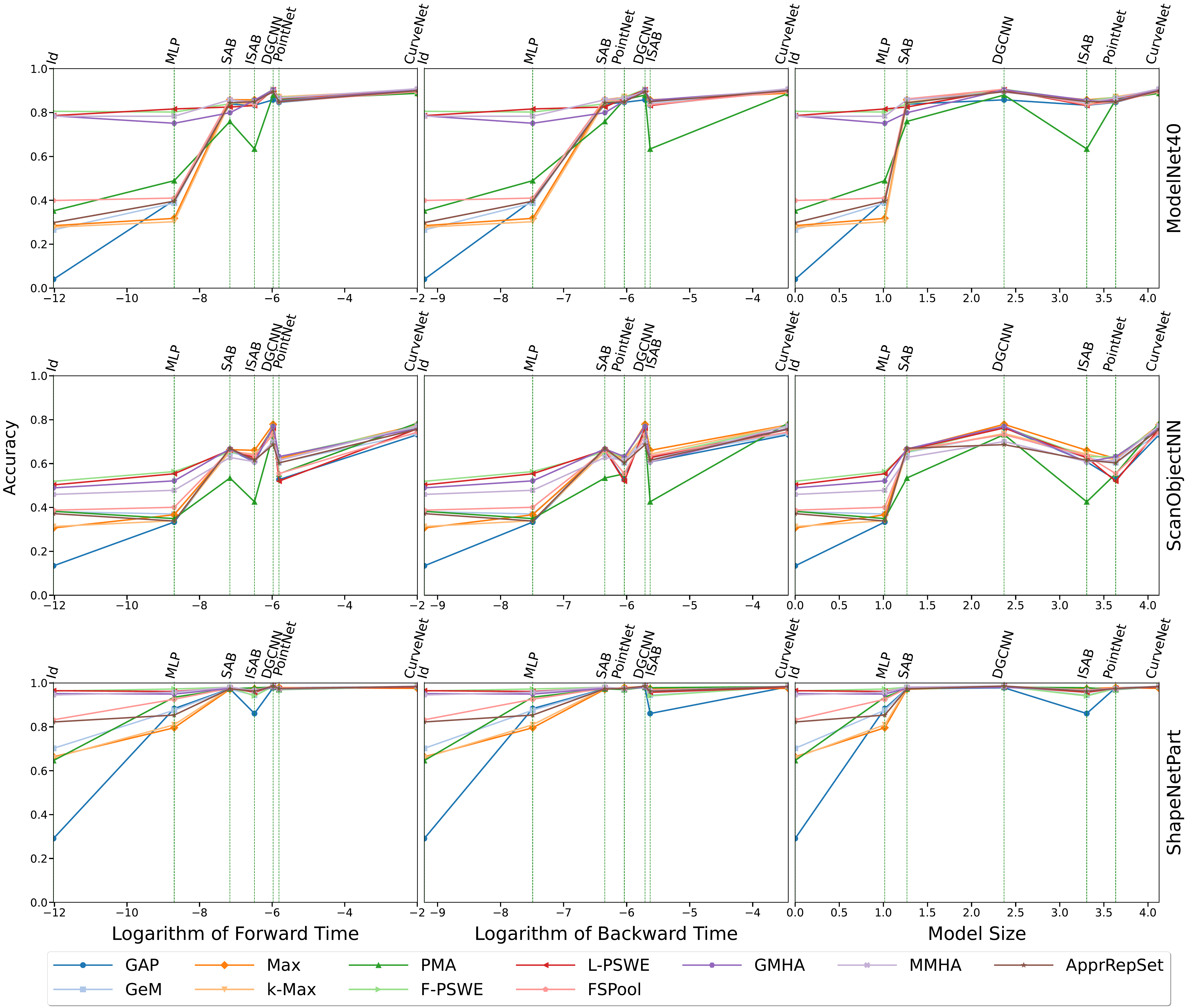}
    \caption{\textbf{Backbone vs. Pooling.} 
     This figure provides a visualization of the models' performances, reported in Table \ref{tab:table(1)}, as a function of the backbone complexity. We employed three indicators as proxies for backbone complexity: average forward time, average backward time, and model size. Each row represents the results from one of the datasets.}  
    \label{fig:bvp_final}
\end{figure*}

To further analyze the findings presented in Table \ref{tab:table(1)}, we examined the performance of pooling layers in relation to the complexity of the backbones. We utilized three indicators as proxies for backbone complexity: average forward time, average backward time, and model size. Figure \ref{fig:bvp_final} displays the performance of different pooling layers for each dataset as a function of the backbones' complexities. Notably, we continue to observe a substantial improvement when utilizing OT-based pooling layers, such as L-PSWE and F-PSWE, and attention-based pooling layers like GMHA and MMHA, in conjunction with the backbones of lower complexities. However, as the complexity of the backbone increases, the advantage conferred by these pooling layers diminishes.

\begin{figure*}[h!]
    \centering    \includegraphics[width=\linewidth]{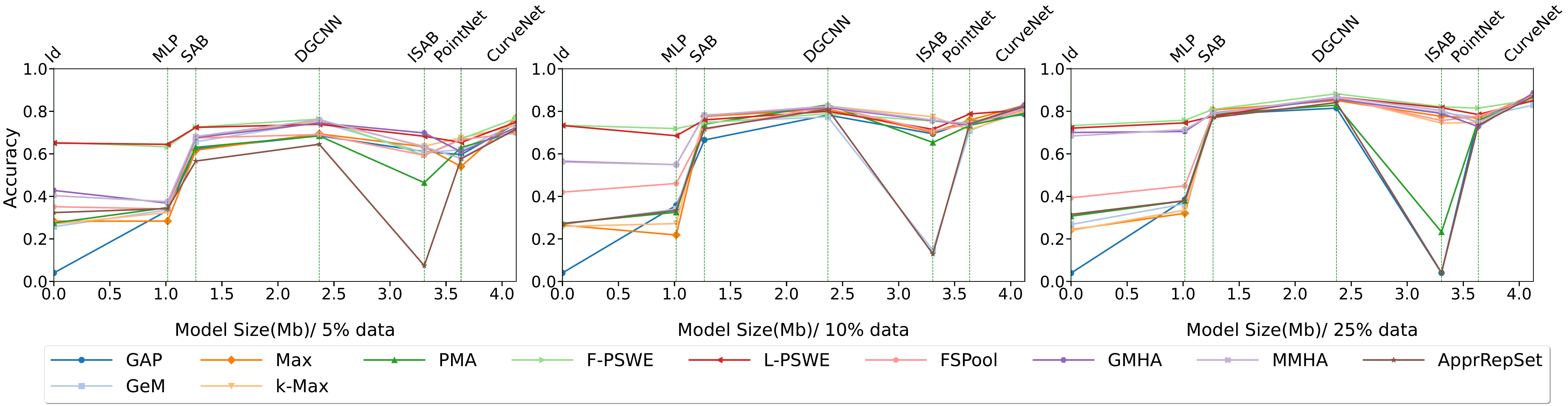}
    \caption{\textbf{Backbone vs. Pooling under Limited Training Data.} The models' performances on ModelNet40, when using 5\% (left), 10\% (middle), and 25\% (right) of the training data. Notably, the OT-based methods show less sensitivity to the sample size.}
    \label{fig:my_label_limited}
\end{figure*}

\textbf{Learning with Limited Training Data.} In our subsequent analysis, we investigated the interplay between pooling methods and backbones when training with limited data. We randomly sampled $5\%$, $10\%$, and $25\%$ of the training data from the ModelNet40 dataset and conducted our experiments accordingly. Figure \ref{fig:my_label_limited} depicts the performance of the models as a function of model size. Notably, we observed the robustness of OT-based pooling methods under limited data conditions. Specifically, for the MLP backbone, L-PSWE and F-PSWE displayed less sensitivity to the training set size, whereas GMHA and MMHA exhibited higher sensitivity. Another significant finding was that while the impact of pooling layers diminished as backbone complexity increased when training on the full dataset, these differences became more pronounced when training with limited data, even for complex backbones.

\begin{figure}[h!]
    \centering    \includegraphics[width=\linewidth]{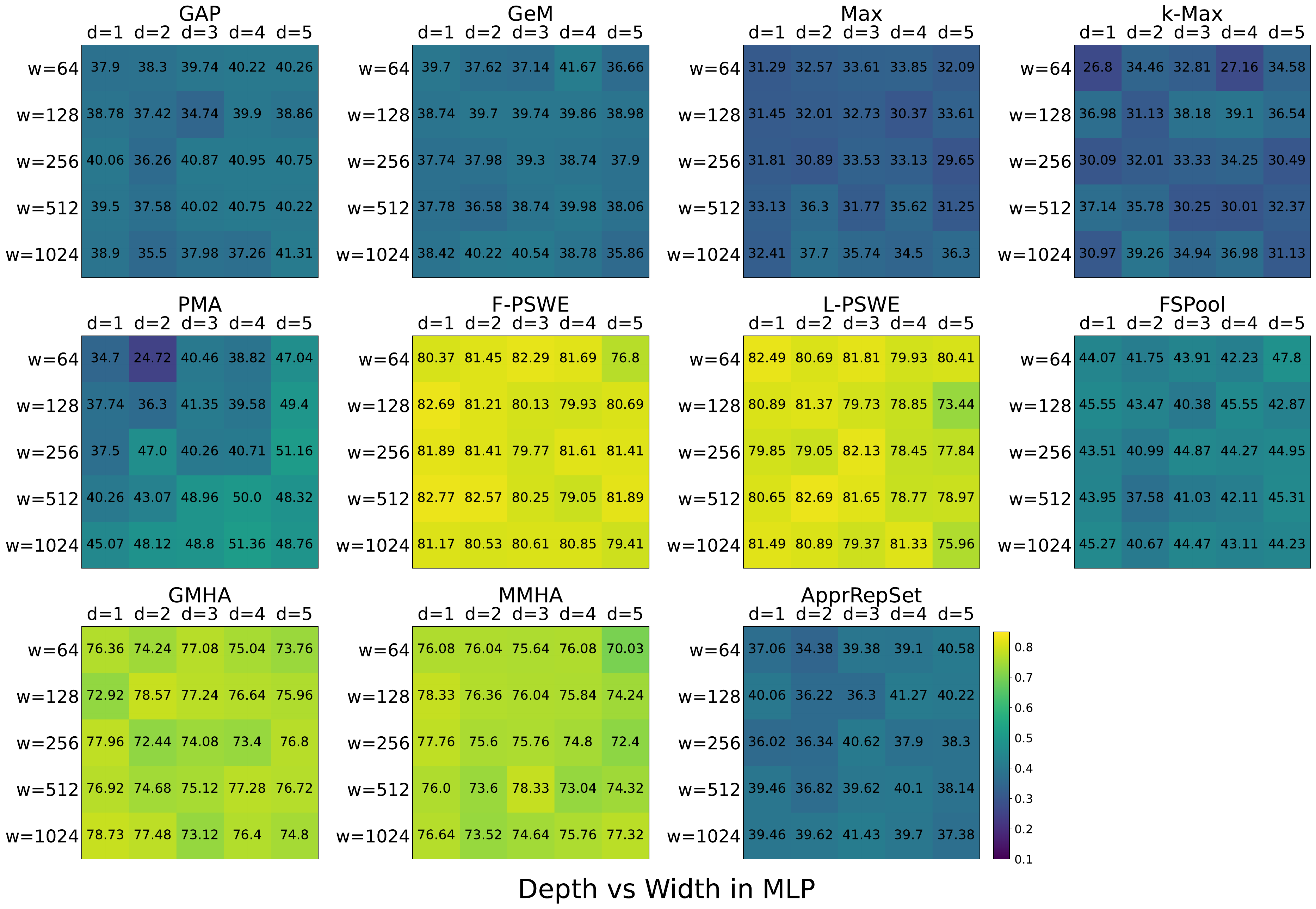}
    \caption{\textbf{Depth vs. Width.} The performance of each pooling method on MLP backbones with varied widths and depths is illustrated in the figure. The results consistently highlight the superior performance of OT-based poolings, particularly F-PSWE and L-PSWE, across different backbone sizes. Additionally, attention-based poolings GMHA and MMHA exhibit competitive performance, closely trailing behind the OT-based methods.}
    \label{fig:my_label_dvsw}
\end{figure}

\textbf{Depth vs. Width.} For models such as MLP and SAB, the width and depth of the model provide control over its size and complexity. It is natural to inquire about the impact of the pooling layer in relation to the backbone size, given a fixed architecture type. To address this, we constructed models with various depths (ranging from 1 to 5) and widths (64, 128, 512, and 1024) for a fixed architecture type. This resulted in 25 different backbones, which, combined with the 11 pooling layers and three repetitions, yielded a total of 825 models per architecture type. We conducted this experiment on the ModelNet40 dataset and presented the results for MLP in Figure \ref{fig:my_label_dvsw}. Once again, we observe the overall superiority of OT-based pooling methods, L-PSWE and F-PSWE, with attention-based approaches such as GMHA and MMHA following closely behind. Similar results for SAB are provided in Appendix \ref{sec:dvw_sab}.

Another notable observation is that while adjusting the width and depth of the backbone impacts the performance to some extent, the most significant performance improvement is achieved through the utilization of better permutation invariant pooling techniques, particularly for DeepSets and Set Transformers.

\textbf{Pairing Pooling Layers.} Combining different global pooling methods, such as GAP and Max pooling, for point cloud classification is a common approach used across a variety of models \cite{dgcnn, Xiang_2021_ICCV, Qi_2017_CVPR, NIPS2017_d8bf84be, goyal2021revisiting, Su_2015_ICCV}. This is done by directly concatenating the output of each pooling method prior to the classification head. Here, we seek to understand which pairings of the pooling layers are complementary.

\begin{figure}[h!]
    \centering    \includegraphics[width=\linewidth]{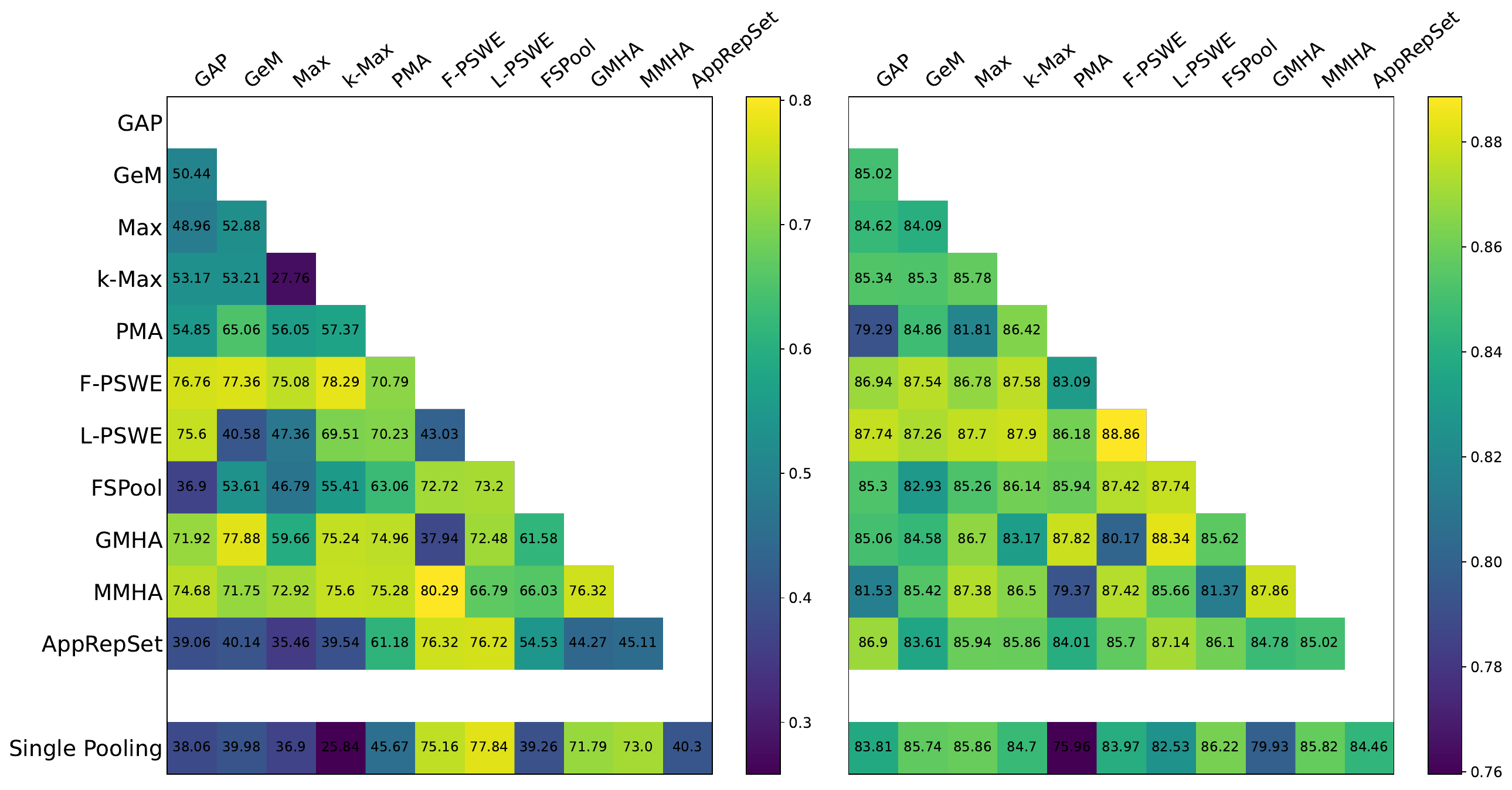}
    \caption{\textbf{Effect of Paired Pooling Methods.} Test set classification accuracy for models trained with the same backbone and different pairs of pooling layers. Results are shown for MLP (left) and SAB (right) backbones. The bottom rows in both plots represent the performance of the same model with a single pooling layer.}
    \label{fig:my_label_coupled_mn}
\end{figure}

We evaluated pairs of pooling layers for fixed backbones on the three datasets, resulting in 55 models per backbone and per dataset. Specifically, we used MLP and SAB as our backbones. Figure \ref{fig:my_label_coupled_mn} displays the performance of models with different pairs of pooling layers for the MLP and the SAB backbones on the ModelNet40 dataset. Additionally, we provide the performance of each individual pooling layer in the bottom row of each plot for the sake of comparison. As expected, we can see that not all pairings are complementary, while many are. In particular, the performance of the F-PSWE pooling method with MLP backbone, $75.16\%$, can be boosted to $80.29\%$ when complemented with MMHA. We also repeated this experiment on the ScanObjectNN and ShapeNet datasets, and the results can be found in Appendix \ref{sec:paired_appendix}.

\clearpage

\section{Discussions and Concluding Remarks}

Point-based methods for deep point cloud classification rely on permutation invariant neural networks, which typically consist of a permutation equivariant backbone, global permutation invariant pooling, and a shallow classifier. Recent models have achieved improved accuracies on benchmark datasets. However, it has been noted that the observed progress may not solely stem from better network architecture designs but also from various auxiliary factors such as evaluation schemes, data augmentation strategies, and loss functions, making it challenging to attribute improvements solely to the network architecture itself \cite{goyal2021revisiting}. Our work explores the intricate interplay between backbone architecture and pooling approaches in point cloud classification to achieve better network architectures. We aim to understand how these factors impact model performance in a controlled setting where training and evaluation conditions are standardized across all models. This allows us to disentangle the effects of network architecture from other factors and gain deeper insights into the relationship between backbone architecture, pooling methods, and classification performance in point cloud analysis.

We trained more than 4,000 models on the three benchmark datasets -- ModelNet40, ScanObjectNN, and ShapeNetPart -- to answer the following questions: How important is the effect of the backbone versus the pooling layer in point cloud classification? How robust are different models when learning with limited data? What is the effect of width and depth of the permutation equivariant backbones for different pooling layers in DeepSets and Set Transformers? And lastly, would pairing permutation invariant pooling layers affect the models' performances? Our findings highlight the performance gap between traditional and advanced pooling methods, the robustness of models under limited data, and the critical role of pooling layers in DeepSets and Set Transformers. Further, we observed that pairing pooling layers could lead to enhanced model performance. 

Our results consistently showed that OT-based pooling methods, such as F-PSWE and L-PSWE, outperform traditional pooling techniques across various backbones and datasets. The superior performance of OT-based methods can likely be attributed to their ability to capture complex relationships and distributions within the feature space, providing a richer and more informative representation of the point cloud data. This suggests that for applications requiring high accuracy and robustness, incorporating OT-based pooling layers is highly beneficial. Meanwhile, attention-based pooling methods, including GMHA and MMHA, also demonstrated strong performance. These methods dynamically assign weights to different features based on their relevance, effectively focusing on the most informative parts of the point cloud. This dynamic weighting mechanism allows attention-based pooling to adaptively aggregate important features, which is particularly useful in scenarios with complex or noisy data.

We also observed that the performance benefits of complex pooling methods are more pronounced with simpler backbones. For instance, when using only Identity as our backbone, advanced pooling methods like F-PSWE and GMHA provide significant performance improvements. However, as the complexity of the backbone increases, the relative gains from these sophisticated pooling methods diminish. This indicates that while complex backbones can learn rich features independently, simpler backbones can substantially benefit from advanced pooling techniques to enhance their representational power.

Our work offers many potential insights for practitioners tackling point cloud classification tasks. First, we find that combining different pooling methods can yield significant performance improvements. This complementary effect arises because different pooling methods capture different aspects of the feature distribution, leading to a more comprehensive aggregation of the point cloud data. Therefore, for practitioners, experimenting with combinations of pooling layers may yield worthwhile performance improvements.  In scenarios with limited training data, OT-based pooling methods exhibit less sensitivity to the sample size compared to alternative pooling methods. This robustness makes OT-based pooling particularly suitable for applications where data is scarce or expensive to collect. Finally, we observe that while both the complexity of the backbone and the pooling layers contribute to overall model performance, it is crucial to strike a balance between these components. For simpler backbones, investing in more sophisticated pooling methods can lead to substantial gains. Conversely, for complex backbones, simpler pooling methods may suffice, allowing computational resources to be allocated elsewhere in the model.

Ultimately, point cloud classification remains a significant area of interest in the machine learning community, with continuous advancements being made. Several works which we did not originally consider in our experiments, such as \cite{wu2022point, ma2022rethinking, chen2023pointgpt, qi2024shapellm}, have presented novel state-of-the-art point cloud classification approaches that further enhance classification accuracy and robustness. Integrating these newer architectures into our experimental framework is thus an important direction for future work. Moreover, while the focus of this work is on permutation invariant model architectures, we additionally note the $SO(3)$ symmetry inherent to point cloud data. This symmetry motivates the use of methods that are invariant to rotation as well. While the use of rotation augmentation in our experimental setup aids in partially addressing this symmetry, recent works such as \cite{Chen_2022_CVPR} and \cite{WANG2024110624} offer promising results by explicitly incorporating rotation invariance into their model architectures. The integration of such methods into our work to achieve a more holistic understanding of point cloud classification approaches presents yet another valuable avenue for future research.

\section*{Acknowledgements}
This research was supported by the NSF CAREER Award
No. 2339898. The authors would like to thank the GRaM workshop reviewers for their valuable feedback, which helped us improve the paper.

\clearpage

\bibliography{refs}
\bibliographystyle{unsrt}

\clearpage
\appendix
\onecolumn

\section{Point Cloud Classification Approaches}
\label{Appendix A: Appendix A}

\subsection{Blueprint of permutation invariant neural networks}

In order to ensure that a function is $\mathcal{G}$-invariant, where $\mathcal{G}$ is a symmetry group, a consistent blueprint can be followed.

\paragraph{Symmetry group} A group $G$ is defined with an operator $\circ: G\times G\rightarrow G$ if it satisfies associativity, closure under $\circ$, has an identity element $e$ for $\circ$, and an inverse for each element $g\in G$. A symmetry group represents symmetries of a geometric object. In this work, we focus on symmetry groups of all permutations of a finite set, particularly in the context of point clouds. We denote by $\mathcal{G}$ the group of all permutations of the points in the point cloud.

\paragraph{$\mathcal{G}$-equivariant and $\mathcal{G}$-invariant} Given a signal $x(\omega)$ on a domain $\omega$ and a function $f$ defined on $x$, $f$ is $\mathcal{G}$-equivariant if it satisfies $f(\rho(g)x(\omega))=\rho(g)f(x(\omega))$ for all $g\in\mathcal{G}$, where $\rho(g)$ is a geometric transformation belonging to $\mathcal{G}$. $f$ is $\mathcal{G}$-invariant if it satisfies $f(\rho(g)x(\omega))=f(x(\omega))$ for all $g\in\mathcal{G}$, meaning $f$ is invariant to geometric transformations in $\mathcal{G}$.

\paragraph{General blueprint for permutation invariant neural networks} To create a permutation invariant network, we can construct a series of permutation equivariant layers followed by a global permutation invariant layer. A function $f$ is considered permutation invariant if it satisfies $f(PX,PAP^T) = f(X,A)$, where $X \in \mathbb{R}^{N\times d}$ is a series of $d$-dimensional input vectors, $P \in \mathbb{R}^{N\times N}$ is a permutation matrix, and $A \in \mathbb{R}^{N\times N}$ is an adjacency matrix. We can express $f$ as follows:
\begin{equation}
\begin{gathered}
f = L_{I} \circ L_{e_m} \circ L_{e_{m-1}} \circ \dots \circ L_{e_1}
\end{gathered}
\end{equation}
where $L_{I}$ is a global permutation invariant layer, which can be a pooling layer or any other type of permutation invariant layer. $\{L_{e_i}\}^{m}_{i=1}$ are permutation equivariant layers, and this blueprint is sufficient to make $f$ invariant to permutation. Additional layers, such as local pooling layers, can be used to address the curse of dimensionality, as long as the entire composed function remains invariant with respect to input permutation.

\subsection{Message passing} 

The message passing mechanism embeds node features in each layer by communicating with their neighbors. Let $G=(V, E, X)$ be a graph with nodes $V=\{v_1, v_2, \dots, v_N\}$, edges $E \subseteq V\times V$, and feature matrix $X=\{x_1,x_2,\dots, x_N\}$, where $x_i\in \mathbb{R}^d$ is the feature vector corresponding to node $v_i$. The adjacency matrix $A^{N\times N}$ denotes the presence of edges in the graph, where $[A]_{i,j} \in \{0,1\}$.

Let $h_i^{(k)}$ be the hidden state of node $i$ at hidden layer $k$, where $k=0, \dots, K$ and $h_i^{(0)} = x_i$. The message passed from node $j$ to node $i$ at iteration $k$ is given by the learnable function $\text{Message}_{\theta}(\cdot, \cdot)$, which takes as input the hidden states of source node $j$ and target node $i$:
\begin{equation}\label{eq:msg}
m_{j\to i}^{(k)} = \text{Message}_{\theta}\left(h_j^{(k-1)}, h_i^{(k-1)}\right).
\end{equation}
The aggregated message for node $i$ is then computed by an \text{Aggregation} function from its neighboring nodes $\mathcal{N}_i$:
\begin{equation}\label{eq:agg}
    m_i^{(k)} = \text{Aggregation}\left(m_{j\to i}^{(k)}\right)_{j\in\mathcal{N}_i}.
\end{equation}
\text{Aggregation} must be a locally permutation invariant operator, e.g., global average or summation. Finally, the updated hidden state of node $i$ is obtained by applying the learnable function $\text{Update}_{\theta}(\cdot, \cdot)$ to the previous hidden state and the aggregated message:
\begin{equation}\label{eq:update}
h_i^{(k)} = \text{Update}_{\theta}\left(h_i^{(k-1)}, m_i^{(k)}\right).
\end{equation}

This mechanism learns helpful representations at the last layer, $K$, and uses them for any downstream tasks.

\subsection{Learning from sets}

Sets are unordered collections of feature vectors, e.g, point clouds and LiDAR scans. Learning from sets is akin to learning from graph-structured data, as sets can be considered as edge-less graphs or fully connected graphs. In the former, each node performs the message passing mechanism on itself, while in the latter, all nodes communicate with each other. This core principle also underlies transformer networks, where similarity between queries and keys determines the attention weights that control information flow.

Considering sets as a fully-connected graph, the message passing is the same as in equations \ref{eq:msg}, \ref{eq:agg}, and \ref{eq:update}. However, considering them as edge-less nodes of a graph results in no communication between nodes. As a result, equation \ref{eq:msg} will become $m_{i}^{(k)} = \text{Message}_{\theta}\left(h_i^{(k-1)}\right)$. Consequently, the aggregated message will be calculated as $m_i^{(k)} = \text{Aggregation}\left(m_{i}^{(k)}\right)$, and the hidden state will be updated only by the node itself.

\section{Backbone and Pooling Hyperparameters}\label{app:hyperparams}



\subsection{Backbones}\label{appendix:backbone}

For the MLP backbone, representing the DeepSets architecture, we use 2 hidden layers of 512 neurons and an output layer of dimension 3. We use Leaky ReLU activation on the hidden layers.

For both the SAB and ISAB backbones in Backbone vs Pooling experiment, we use 1 hidden layer of dimension 256 and 4 heads. Furthermore, for the ISAB backbone, we use 16 inducing points in the architecture.

We use the original backbone implementation without modification of DGCNN, PointNet, and CurveNet, as published by the respective authors of each backbone.

\subsection{Poolings}\label{appendix:pool}

For the Generalized Mean pooling, we use a moment of 2. For the k-Max pooling, we use $k=2$. For PMA we use 1024 seed vectors with just 1 head. For both L-PSWE and F-PSWE we use 1024 reference points and 1024 projections. For FSPool, we use 1024 pieces in the piecewise linear function and relaxed sorting. 

For both GMHA and MMHA, we use 4 heads, while we use 2 affine layers for GMHA and 1 affine layer for MMHA. We do not use the temperature hyperparameter.

Finally, for ApprRepSet, we use 3 hidden sets with 5 elements.

\clearpage
\section{Depth vs. Width Results for SAB Backbone}\label{sec:dvw_sab}

We observed that the SAB backbone, which we consider to be a more ``complex'' backbone, performs well in general, regardless of the pooling method in use. This is consistent with previously observed results, which showed that the benefit of particular pooling layers, such as the OT-based and attention-based poolings, diminished with increasing backbone complexity. Furthermore, as expected, increasing the depth and width beyond certain limits leads to a decrease in the test accuracy.

\begin{figure}[H]
    \centering    \includegraphics[width=\linewidth]{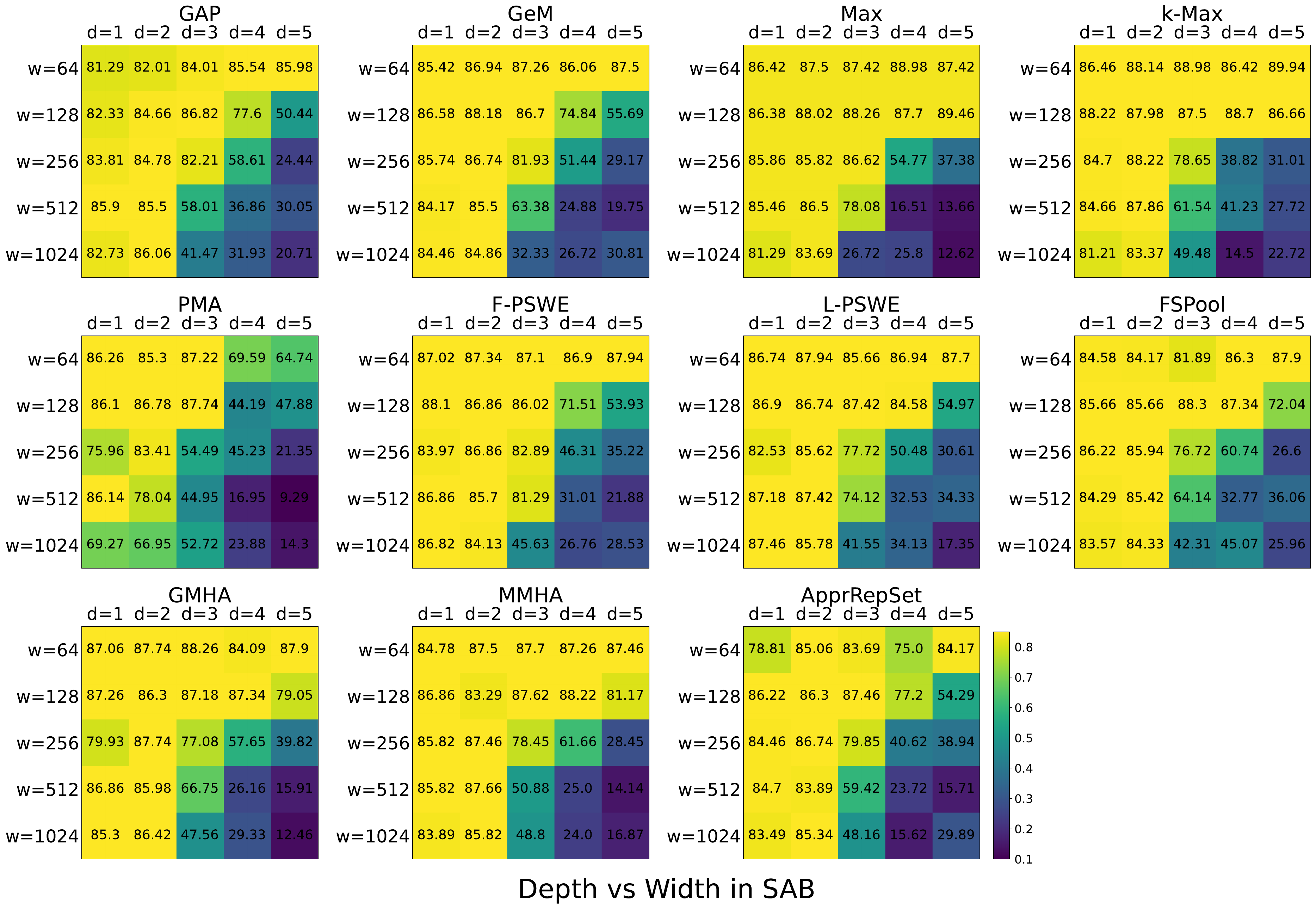}
    \caption{\textbf{Depth vs. Width.} The performance of each pooling method on SAB backbones with varied widths and depths is illustrated in the figure.}
    \label{fig:my_label_dvsw_sab}
\end{figure}

\clearpage
\section{Paired Pooling Results on ScanObjectNN and ShapeNetPart}\label{sec:paired_appendix}

\subsection{ScanObjectNN}
\begin{figure}[h!]
    \centering    \includegraphics[width=\linewidth]{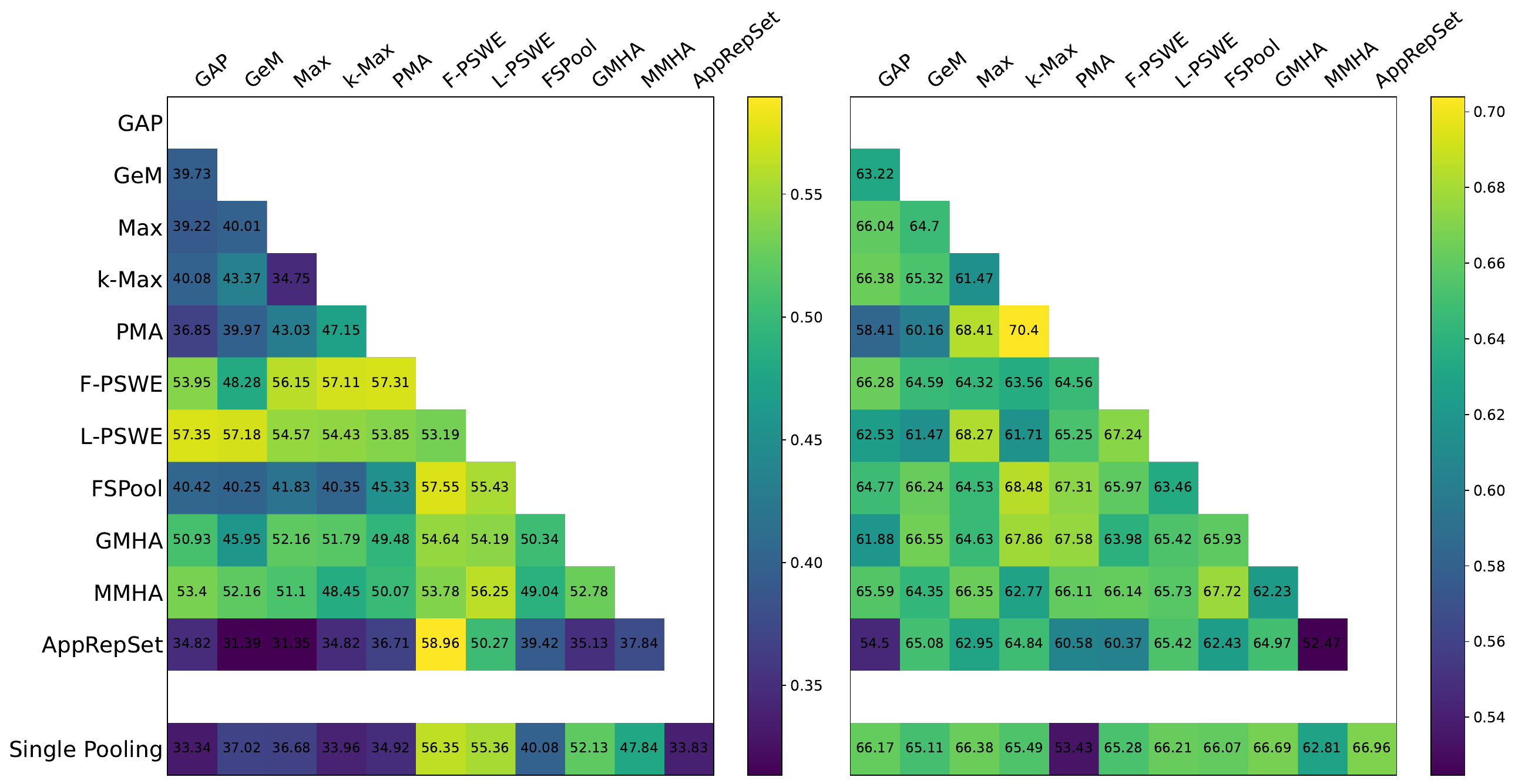}
    \caption{\textbf{Effect of Paired Pooling Methods on ScanObjectNN.} Test set classification accuracy for models trained with the same backbone and different pairs of pooling layers. Results are shown for MLP (left) and SAB (right) backbones. The bottom rows in both plots represent the performance of the same model with a single pooling layer.}
    \label{fig:pair_scan}
\end{figure}

\clearpage

\subsection{ShapeNetPart}
\begin{figure}[h!]
    \centering    \includegraphics[width=\linewidth]{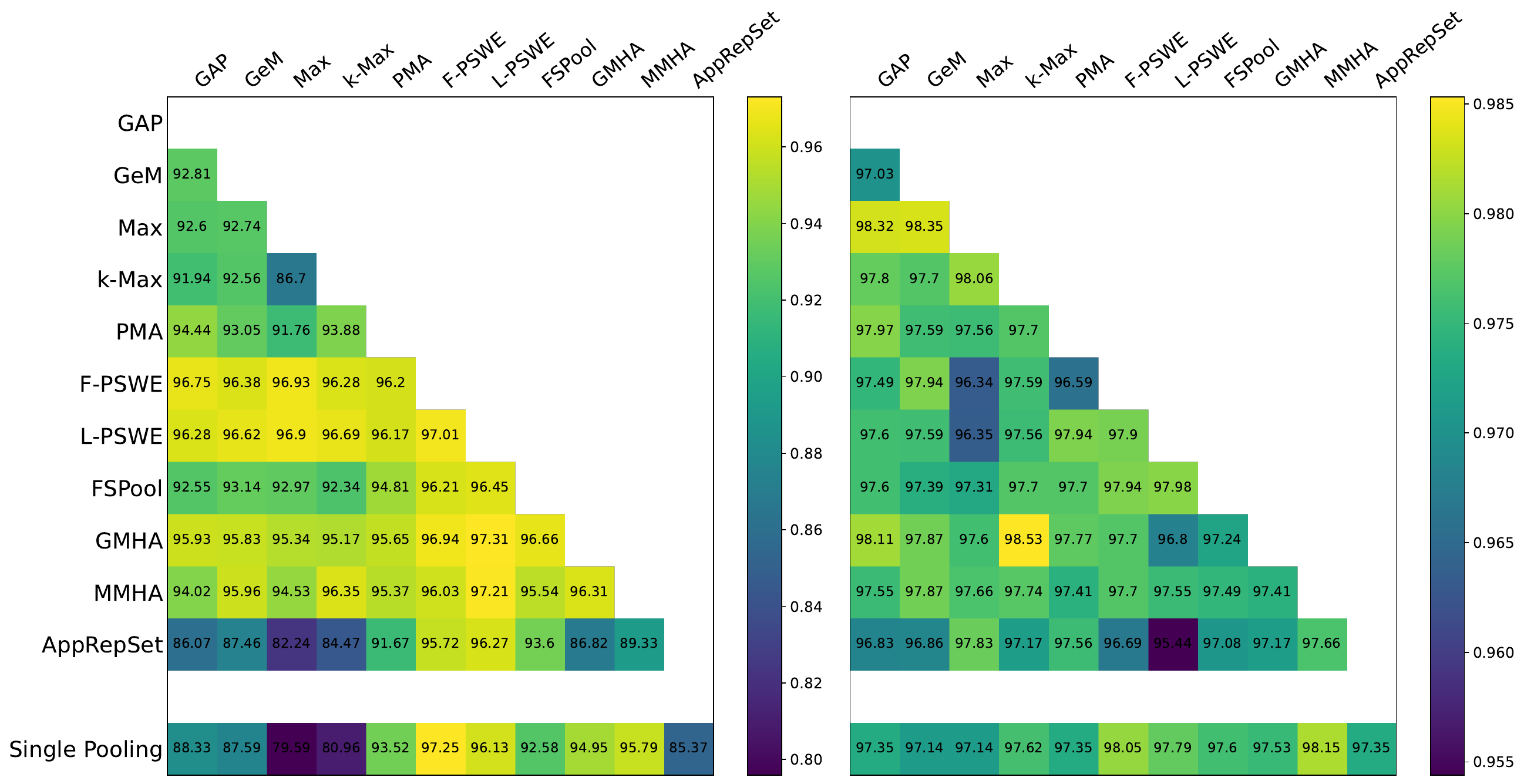}
    \caption{\textbf{Effect of Paired Pooling Methods on ShapeNetPart.} Test set classification accuracy for models trained with the same backbone and different pairs of pooling layers. Results are shown for MLP (left) and SAB (right) backbones. The bottom rows in both plots represent the performance of the same model with a single pooling layer.}
    \label{fig:pair_shape}
\end{figure}

As demonstrated in Figures \ref{fig:pair_scan} and \ref{fig:pair_shape}, combining specific pooling methods can enhance performance, though we once again observe that not all poolings are compatible. In the case of ScanObjectNN, we find that, while the OT-based and attention-based poolings generally perform the best in the single pooling case, pairing k-Max pooling with PMA pooling boosts test accuracy to $70.4\%$ with the SAB backbone. When considering ShapeNetPart, we observe that pairing L-PSWE with the attention-based GMHA and MMHA poolings can result in a performance boost with the MLP backbone. These experiments highlight that pairing certain complementary poolings at the final permutation invariant layer can lead to notable performance improvements.

\end{document}